\documentclass[10pt,twocolumn,letterpaper]{article}

\usepackage{iccv}
\usepackage{times}
\usepackage{epsfig}
\usepackage{graphicx}
\usepackage{amsmath}
\usepackage{amssymb}
\usepackage {appendix}
 
\usepackage{multirow}
\usepackage{caption}
\usepackage{lipsum}
\usepackage{bm}
\usepackage{subcaption}
\usepackage{xcolor}
\definecolor{demphcolor}{RGB}{144,144,144}
\newcommand{\demph}[1]{\textcolor{demphcolor}{#1}}

\usepackage[breaklinks=true,bookmarks=false]{hyperref}

\iccvfinalcopy 

\newcommand\blfootnote[1]{%
  \begingroup
  \renewcommand\thefootnote{}\footnote{#1}%
  \addtocounter{footnote}{-1}%
  \endgroup
}

\def\iccvPaperID{****} 

\ificcvfinal\fi

\begin{document}

\title{Designing a Better Asymmetric VQGAN for StableDiffusion}

\author{Zixin Zhu$^{1*}$
\quad Xuelu Feng$^{1*}$ \quad Dongdong Chen$^2$ \quad Jianmin Bao$^2$ 
\quad Le Wang$^{1}$ \quad Yinpeng Chen$^2$  \\
\quad Lu Yuan$^2$ 
\quad Gang Hua $^{1,3}$\\
{$^1$Xi'an Jiaotong University} \quad  {$^2$Microsoft} \quad {$^3$Wormpex AI Research} \\
\small{\texttt{\{zhuzixin@stu., xueluF@stu., lewang@\}xjtu.edu.cn}}\\
\small{\texttt{\{jianbao, yiche,luyuan\}@microsoft.com}},\quad \small{\texttt{\{cddlyf, ganghua\}@gmail.com}}
}


\twocolumn[{
\renewcommand\twocolumn[1][]{#1}
\maketitle
    \vspace{-1.9em}
    \setlength\tabcolsep{0.5pt}
    \begin{center}
    \captionsetup{type=figure}
    \includegraphics[width=1\linewidth]{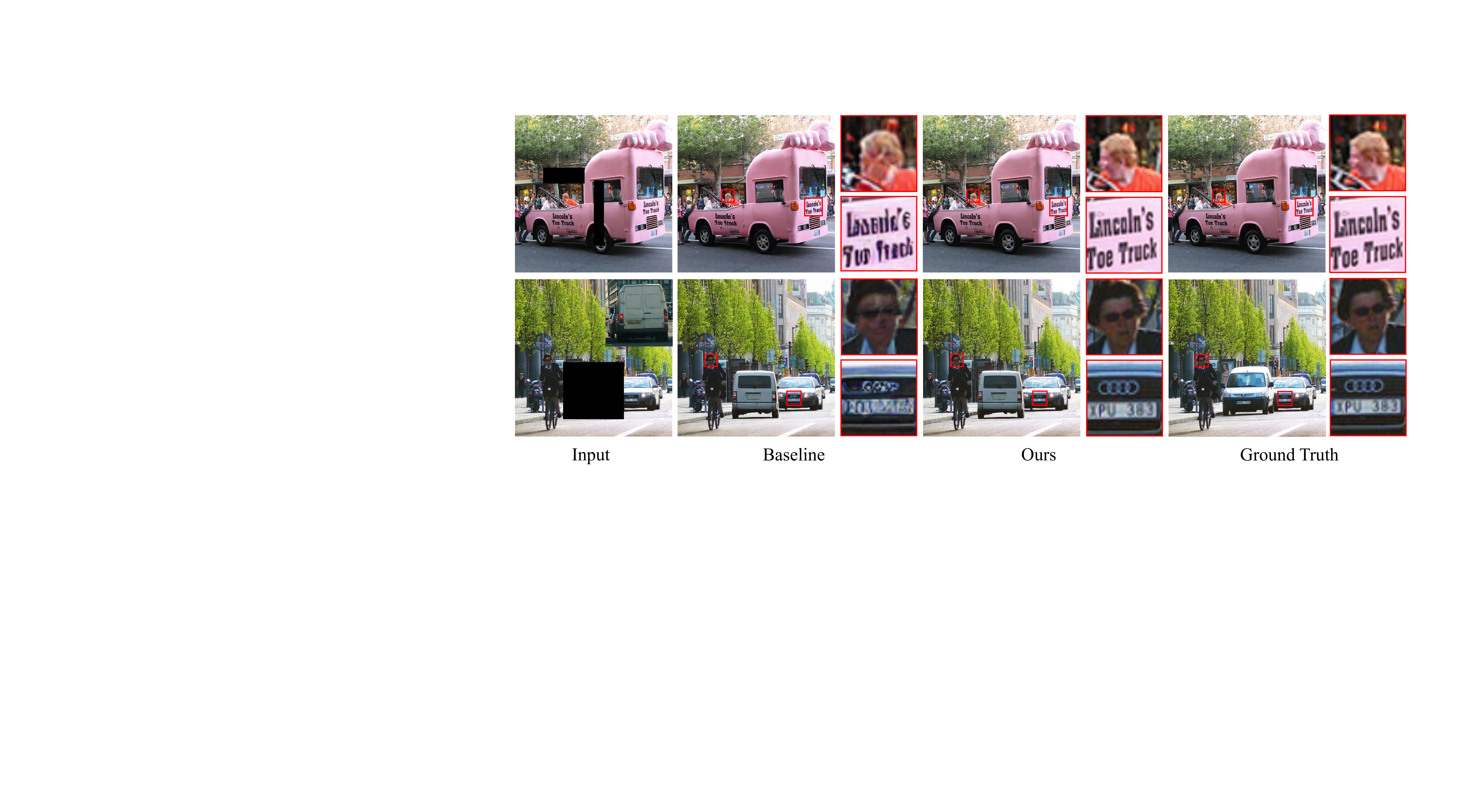}
    \captionof{figure}{Comparing StableDiffusion-based inpainting (top) / editing \cite{yang2022paint} (bottom) results with default VQGAN and our proposed asymmetric VQGAN. The vanilla VQGAN in StableDiffusion will cause serious distortion even for the non-edited regions. In contrast, our proposed asymmetric VQGAN can preserve more details and deliver superior results.}
    \label{fig:motivation}
\end{center}
}]

\maketitle

\blfootnote{\textsuperscript{*} Work done during internship at Microsoft}
\ificcvfinal\thispagestyle{empty}\fi

\begin{abstract}
StableDiffusion is a revolutionary text-to-image generator that is causing a stir in the world of image generation and editing. Unlike traditional methods that learn a diffusion model in pixel space, StableDiffusion learns a diffusion model in the latent space via a VQGAN, ensuring both efficiency and quality. It not only supports image generation tasks, but also enables image editing for real images, such as image inpainting and local editing. However, we have observed that the vanilla VQGAN used in StableDiffusion leads to significant information loss, causing distortion artifacts even in non-edited image regions. To this end, we propose a new asymmetric VQGAN with two simple designs. Firstly, in addition to the input from the encoder, the decoder contains a conditional branch that incorporates information from task-specific priors, such as the unmasked image region in inpainting. Secondly, the decoder is much heavier than the encoder, allowing for more detailed recovery while only slightly increasing the total inference cost. The training cost of our asymmetric VQGAN is cheap, and we only need to retrain a new asymmetric decoder while keeping the vanilla VQGAN encoder and StableDiffusion unchanged. Our asymmetric VQGAN can be widely used in StableDiffusion-based inpainting and local editing methods. Extensive experiments demonstrate that it can significantly improve the inpainting and editing performance, while maintaining the original text-to-image capability. The code is available at \url{https://github.com/buxiangzhiren/Asymmetric_VQGAN}.
  
\end{abstract}

\section{Introduction}

Diffusion models have emerged as the most popular generative models, achieving remarkable results in image synthesis. Early diffusion models required significant computational resources, as they performed the diffusion process in the high-dimensional pixel space of RGB images. To reduce the training cost while preserving the generation quality, laten diffusion model (LDM)~\cite{rombach2022high} employs VQGAN~\cite{esser2021taming} to move the diffusion step to a low-dimensional latent space. In a subsequent development, StableDiffusion has further scaled up LDM with a larger model and data scale, resulting in a highly powerful general text-to-image generator. Since its public release, it has drawn significant attention in the world of generative AI.

StableDiffusion not only possesses text-to-image generation capabilities, but it also supports various editing-related tasks, such as inpainting \cite{rombach2022high} and local editing tasks \cite{brooks2022instructpix2pix, mokady2022null, yang2022paint}. For these editing tasks, StableDiffusion can generate new content for selected regions based on user-supplied input condition while aiming to preserve other regions. However, we have observed that the results of StableDiffusion-based editing in all existing methods \cite{rombach2022high, yang2022paint} suffer from distortion artifacts in the non-edited image regions, especially for the regions with fine-grained structures (e.g., texts). For example, as depicted in Figure \ref{fig:motivation}, despite our intention to only inpaint the black mask region or composite the object provided in the reference image into the mask region, we observe severe distortion in the non-mask/non-edited areas.

After extensive analysis, we have found that these issues are caused by the quantization error present in the vanilla VQGAN utilized by StableDiffusion. Specifically, VQGAN utilizes an encoder to downsample images multiple times into a latent space, after which the downsampled image vectors are quantified based on a codebook. As a result, quantization errors are inevitable even for the non-edited regions if we only feed the output of the encoder into the decoder, as the vanilla VQGAN operates by default. Additionally, during the inference process, the convolutional layers utilized in VQGAN's encoder impact the feature vectors of non-masked regions due to the masked regions.

To this end, we propose a new asymmetric VQGAN with two simple yet effective designs in the decoder part. Firstly, we reformulate the VQGAN decoder as a conditional decoder to better support local editing tasks. This is achieved by incorporating an extra branch that can integrate information from task-specific priors. For instance, for inpainting or local editing tasks, we feed the non-edited regions into this branch so that the decoder can use both the output of the encoder and the original non-edited regions as inputs. In contrast, the vanilla VQGAN decoder only takes the output of the VQGAN encoder as input. Secondly, we enhance the capability of the decoder by using deeper or wider decoders rather than similar complexity as the encoder. This stronger decoder can better preserve the non-edited regions and recover more details from the quantized output of the encoder. Considering that the most time-consuming part of StableDiffusion inference is the iterative diffusion process, this larger decoder only slightly increases the total inference cost.

In addition to the inference cost, the training cost of our asymmetric VQGAN is still very cheap. We only need to retrain a new asymmetric decoder while keeping StableDiffusion and the original vanilla VQGAN encoder unchanged. Additionally, by alternatively feeding/not feeding the task-specific priors into the decoder, our asymmetric VQGAN can naturally support both editing tasks that require the task-specific priors and pure generation tasks such as text-to-image generation, which do not require such prior input.

To demonstrate the effectiveness of our asymmetric VQGAN, we conducted experiments on three different tasks. In inpainting and local editing tasks with masks (paint-by-example \cite{yang2022paint}), our asymmetric VQGAN achieved state-of-the-art performance (1.03 FID on the Place dataset \cite{zhou2017places} and 86.35\% CLIP score on the COCOEE dataset \cite{yang2022paint}). In the pure text-to-image task, our model can also achieve comparable or even better results compared to the original StableDiffusion. Our contributions can be summarized in the below three-folds:

\begin{itemize}
\item To the best of our knowledge, we are the first that explicitly point out and investigate the distortion problem in StableDiffusion-based editing methods.

\item We design a new asymmetric VQGAN to address the above distortion issues with two simple yet effective designs. Compared to the typical symmetric VQGAN design,  this new design can better preserve the non-edited regions and recover details, while maintaining a low training and inference cost.

\item Our asymmetric VQGAN achieves state-of-the-art performance on two representative tasks: the inpainting task on the Place dataset \cite{zhou2017places} and the local editing task (i.e., paint by example \cite{yang2022paint}) on the COCOEE dataset.
\end{itemize}

\section{Related Work}

\subsection{Diffusion Models}
Diffusion models are a powerful family of generative models that have recently evolved and drawn significant attention due to their impressive performance on various tasks. Recent works~\cite{dhariwal2021diffusion,song2020denoising} have demonstrated that diffusion model can achieve astonishing results in high-fidelity image generation, even outperforming generative adversarial networks. Diffusion models are naturally ideal for learning models from complex and diverse data, and many variants have been proposed recently. For instance, Denoising Diffusion Probabilistic Models (DDPMs)~\cite{ho2020denoising} are the most popular diffusion models that learn to perform a diffusion process on a Markov chain of latent variables. And Denoising Diffusion Implicit Models (DDIMs)~\cite{song2020denoising} is further proposed to accelerate the denoising process. To improve the efficiency while preserving high generation quality, Latent Diffusion Models (LDM)~\cite{rombach2022high} propose to learn a diffusion model within the latent space rather than the pixel space.

Diffusion models have proven to be highly effective in a variety of applications, including image generation~\cite{ramesh2022hierarchical,nichol2021glide,ho2022cascaded,batzolis2021conditional}, image-to-image translation~\cite{choi2021ilvr,wang2022semantic,wang2022sindiffusion}, super-resolution~\cite{rombach2022high}, and image editing~\cite{nichol2021glide,avrahami2022blended}. Particularly, recent advancements in diffusion models~\cite{sohl2015deep} have led to state-of-the-art image synthesis~\cite{dhariwal2021diffusion,esser2021taming,esser2021imagebart,nichol2021glide,saharia2022photorealistic,gu2022vector,balaji2022ediffi,rombach2022high} as well as generative models for other modalities such as text~\cite{li2022diffusion}, audio~\cite{kong2020diffwave}, and video~\cite{ho2022video}. In this paper, we focus on designing a new VQGAN architecture that can improve the performance of StableDiffusion like diffusion models that operates in the latent space, for image generation and editing tasks. 

\subsection{VQGAN}
The Vector Quantized Variational Autoencoder (VQ-VAE) ~\cite{peng2021generating,razavi2019generating} is a widely used method for learning discrete image representations. VQ-based techniques have been successfully applied to image generation and completion, leveraging a learned codebook in the feature domain. While ~\cite{peng2021generating} extended this approach to use a hierarchy of learned representations, these methods still rely on density estimation from convolutional operations, making it challenging to capture long-range interactions in high-resolution images. In order to address above issues, the Vector Quantized Generative Adversarial Network (VQGAN) ~\cite{esser2021taming} was proposed, which identified that a powerful latent space auto-encoder is critical to capture image details for the following generation stage. So VQGAN uses the quantization procedure of VQVAE and improves the richness of VQVAE’s learned codebook. VQGAN uses the quantization procedure of VQVAE and improves the richness of its learned codebook. Specifically, it employs adversarial loss and perceptual loss to train a better autoencoder in the first stage for synthesizing greater image details. VQ-based generative models have been applied in many tasks, including text~\cite{li2022diffusion}, audio~\cite{kong2020diffwave}, image inpainting ~\cite{liu2022reduce}, and video~\cite{ho2022video} generations. While VQGAN has numerous benefits, the quantization error it introduces can lead to losing image details and causing serious distortion. Motivated by this observation, unlike the conventional symmetric VQGAN design, we explore a new asymmetric VQGAN to retain more details, which can benefit both image generation and editing tasks.
\begin{figure}
\begin{center}
\includegraphics[width=1\linewidth]{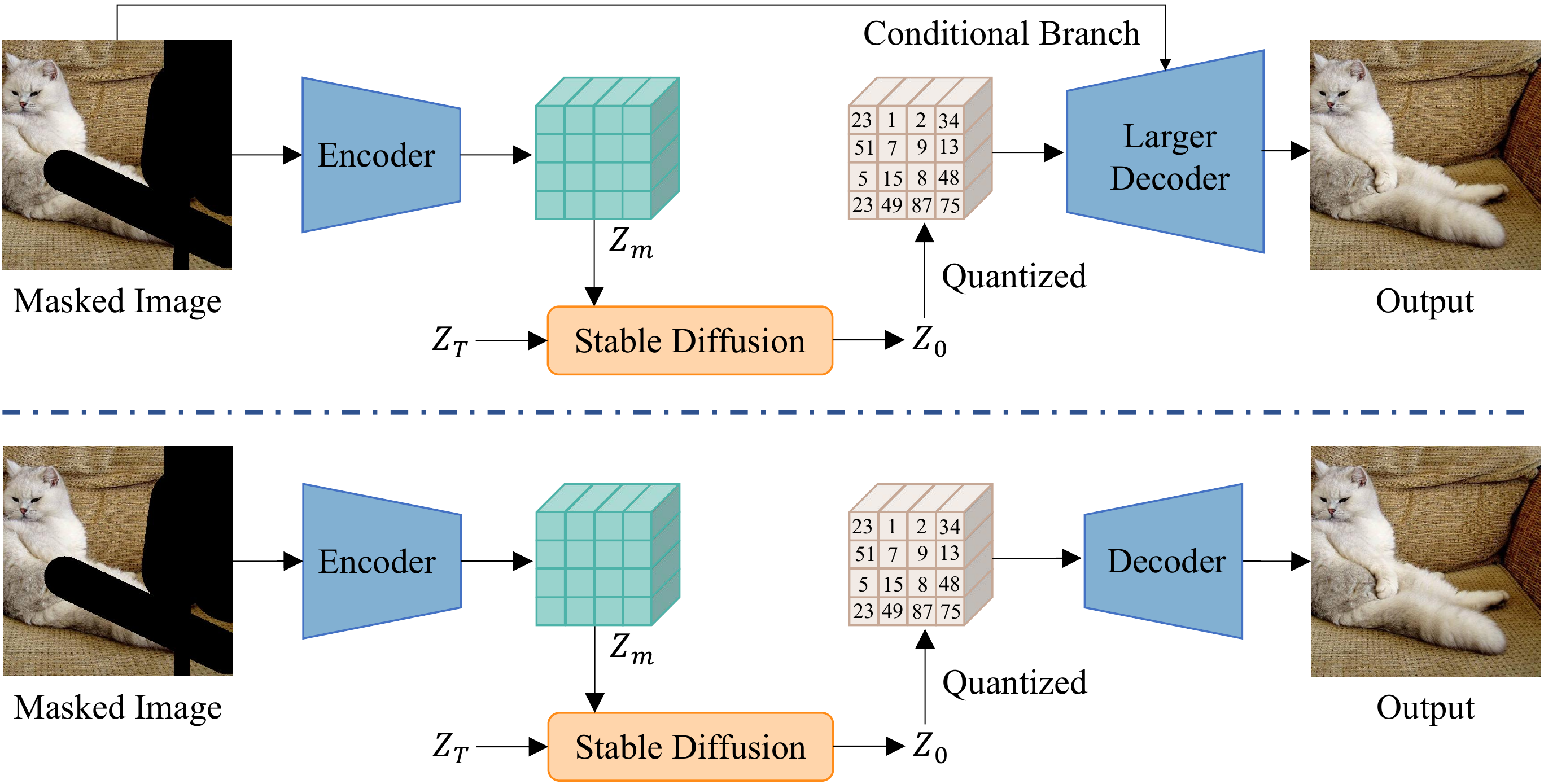}
\end{center}
   \caption{Top: The inference process of our symmetric VQGAN. Bottom: The inference process of vanilla VQGAN.}
\label{fig:framework}
\end{figure}

\begin{figure*}
\begin{center}
\includegraphics[width=1\linewidth]{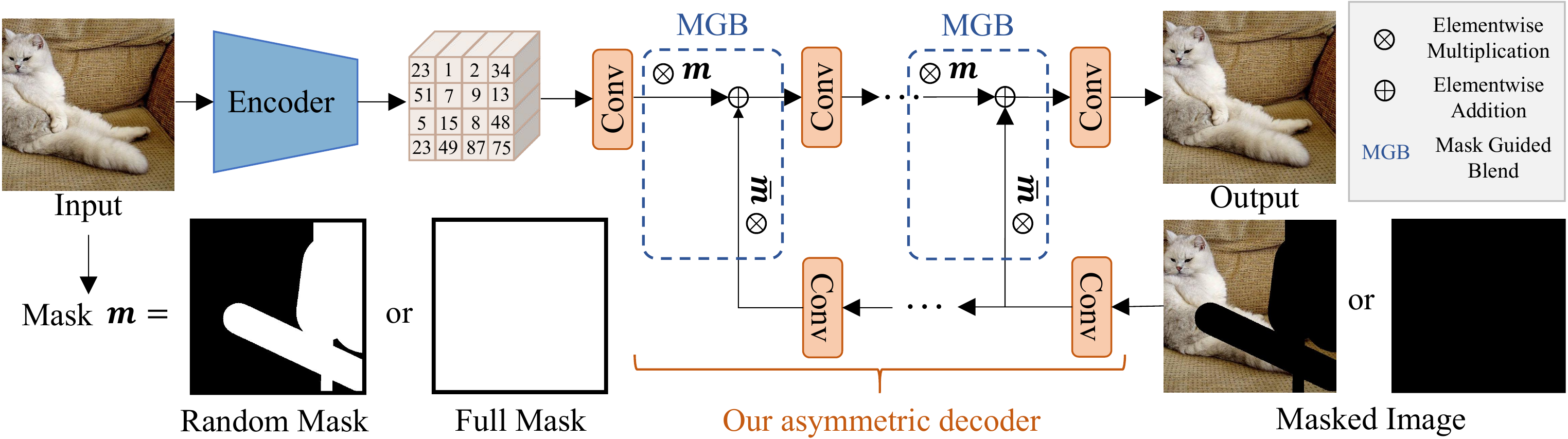}
\end{center}
   \caption{The training process of our symmetric VQGAN. We generate two kinds of masks, \textit{i.e.}, random mask and full mask. The quantized vector from the encoder is fed to our asymmetric decoder. At the same time, the masked image is sent to our decoder as the input of conditional branch. After the conditional and main branches blend, the decoder output final results.}
\label{fig:modeltrain}
\vspace{-2em}
\end{figure*}

\section{Method}

 VQGAN plays an important role in StableDiffusion to map the original high-dimensional pixel space to low-dimensional latent space. However, this mapping process can lead to information loss in image conditional tasks, causing a lack of detail that hurt the quality of the generated result. In this section, we will first discuss the issue of information loss in VQGAN and then introduce our solution, the asymmetric VQGAN, which serves to address this challenge.

\subsection{Information loss in VQGAN}
\label{sec:vqgan}

VQGAN aims to compress the pixel space into discrete latent space. Suppose $\mathbf{X}\in\mathbb{R}^{\textit{H}\times\textit{W}\times3}$ is the input image. VQGAN first utilizes a CNN-based Encoder to obtain its feature variable $\hat{\mathbf{z}} \in \mathbb{R}^{ h \times w \times n_z}$, where $h \times w$ is the spatial resolution and $n_z$ is the channel of the latent vector. Then VQGAN aims to be able to represent it with discrete codebook $\{\mathbf{z}_k\}_{k=1}^K$, where each spatial code $\hat{\mathbf{z}}_{i j}$ find its closest codebook entry $\mathbf{z}_k$ in the codebook, the process can be denoted as follows:
\begin{equation}
\mathbf{z_{i j}}=\mathbf{q}(\hat{\mathbf{z}}_{i j}):=\underset{k \in {1,2,..,K}}{\arg \min }\left\|\hat{\mathbf{z}}_{i j}-\mathbf{z}_k\right\| 
\end{equation}
where $q$ is the quantization encoder that maps the vector to an index of the codebook, Based on the quantized codewords $\mathbf{z}$, VQGAN then adopts the decoder to reconstruct the input image $x$. Suppose the reconstructed result is $\hat{x} = Dec(\mathbf{q}(Enc(\mathbf{x})) $. 

Then the model and codebook can be trained end-to-end via the loss function:
\begin{equation}
\begin{aligned}
\mathcal{L}_{\mathrm{VQ}}(Enc, Dec, \{\mathbf{z}_k\}_{k=1}^K)&=\mathcal{L}_{pixel}  + \lambda \mathcal{L}_{percep} \\& +\left\|\operatorname{sg}[Enc(\mathbf{x})]-\mathbf{z}\right\|_2^2 \\& +\beta\left\|\operatorname{sg}\left[\mathbf{z}\right]-Enc(\mathbf{X})\right\|_2^2 .
\end{aligned}
\end{equation}
where $\mathcal{L}_{pixel}=\|\mathbf{x}-\hat{\mathbf{x}}\|^2$ is pixel-level loss and $\mathcal{L}_{percep}$ perceptual loss calculated with VGG16~\cite{simonyan2014very} network. $\text{sg}[\cdot]$ is the stop-gradient operator, $\beta$ is a hyper-parameter for loss weight. To further improve the quality of the generated samples, a discriminator is employed to perform an adversarial training process with the encoder and decoder. Since the continuous pixel space is mapped into limited discrete space, the information loss phenomenon exists during this process.

We also notice that in some versions of StableDiffusion, the compressing from pixel space to latent space is trained by a KL-reg. The KL-reg shares a similar purpose to VQGAN since they all try to avoid arbitrarily high-variance latent spaces, and both of them share a similar issue: information loss from pixel space to latent space.

\subsection{Asymmetric VQGAN}
Due to the impressive performance of StableDiffusion on text2image generation, it has been widely applied to various conditional image generation tasks. One of the most important and typical of these conditions is the use of image input. However, according to our analysis, these image conditions must be mapped into latent space to satisfy the diffusion process of StableDiffusion. As a result, these conditional images may lose some of their original information during manipulation. Our focus in this paper is to preserve the information of the conditional image input while leaving the pre-trained weights of StableDiffusion unchanged.

To this end, we propose the Asymmetric VQGAN, to preserve the information of conditional image input. Asymmetric VQGAN involves two core designs compared with the original VQGAN as shown in Figure~\ref{fig:framework}. First, we introduce a conditional branch into the decoder of the VQGAN which aims to handle the conditional input for image manipulation tasks. Second, we design a larger decoder for VQGAN to better recover the losing details of the quantized codes. In the next section, we will introduce the detailed structure and training strategy of Asymmetric VQGAN.

\paragraph{Conditional decoder.} We design a conditional decoder that aims to preserve the details of the conditional input. As illustrated in Figure~\ref{fig:modeltrain}, Suppose the conditional image is a masked input $\bm{Y}$ with mask $\bm{m}$, We propose to represent the conditional image input as multi-level feature maps, instead of compressing it into single-layer features. Concretely, we feed the conditional input $\bm{Y}$into a lightweight encoder $E$ and extract the feature map at different layers as the conditional input representation. More formally, we can define
\begin{equation}
\bm{f}_{E}(\bm{Y}) = \left\{\bm{f}_{E}^1(\bm{Y}), \bm{f}_{E}^2(\bm{Y}), \cdots \bm{f}_{E}^n(\bm{Y})\right\},
\label{eqn:conditional_embeding}
\end{equation}
where $\bm{f}_{E}^{k}(\bm{Y})$ represents the $k$-th level feature map from the encoder $E$, $n$ is the number of feature levels. 

Then these features will be integrated into the decoder via a mask-guided blending (MGB) module. MGB aims to preserve the decoders' capability for decoding latent codes while making full use of the features from encoder $E$. It utilizes a mask to directly copy the masked region of the decoder feature while combining the unmasked region feature from the encoder $E$. Specifically, suppose the feature at $k$-th level of the decoder is $\bm{f}_{Dec}^k(\bm{z})$. So the blending process can be formulated as:

\begin{equation}
\bm{f}_{Dec}^k(\bm{z}) = \bm{f}_{Dec}^k(\bm{z}) \otimes \bm{m} + \bm{f}_{E}^k(\bm{Y}) \otimes \bar{\bm{m}},
\end{equation}
where $\otimes$ is element-wise multiplication, $\bar{\bm{m}} = 1 - \bm{m}$. With this designed mask-guided blending module, we do not require any modification to the decoder and just insert several MGB modules into the decoder network while keeping the structures unchanged.

\paragraph{Larger decoder.} To further enhance the capability of decoders for recovering details from given latent codes, we enlarge the decoder model size of the original VQGAN. Increasing the model size of VQGAN is efficient since during the inference stage of StableDiffusion for conditional image input tasks, the decoder only requires to be forwarded one time while the StableDiffusion model needs to be forwarded many times with a significantly larger model size.

\paragraph{Training strategies.} During training, we use the same weights and codebook from the original VQGAN in the Asymmetric VQGAN and only train the new decoder. To avoid the decoder from developing a simple solution of recovering information only from the conditional masked input, we consider two scenarios: one where the mask is randomly generated, and one where the mask is completely filled with 1, meaning the decoder needs to rely on the latent codes to recover the image. We alternately use these two scenarios for 50\% of the training process. For the training objectives, we use the pixel level loss, perceptual loss, and adversarial loss as described in Section~\ref{sec:vqgan} to only update the weights of the Decoder.

Asymmetric VQGAN is lightweight and flexible, it keeps the encode and latent space diffusion process unchanged, in which we can leverage all the pre-trained weights of the Encoder and StableDiffusion. We just need to change the decoder part while enjoying the strong capability of StableDiffusion for a wide range of tasks.

\begin{table}

\begin{center}
\scalebox{0.87}{
\begin{tabular}{l c c c c c}
\hline 
\multirow{2}{*}{Method} & \multicolumn{2}{c}{40-50\% masked} & & \multicolumn{2}{c}{All samples}\\
\cline{2-3} \cline{5-6}
\multirow{2}{*}{} & FID$\downarrow$ & LPIPS$\downarrow$ & & FID$\downarrow$ & LPIPS$\downarrow$ \\
\hline\hline
LaMa~\cite{suvorov2022resolution} & 12.0 & 0.24 & & 2.21 & 0.14\\
CoModGAN~\cite{zhao2021large} & 10.4 & 0.26 & &1.82 & 0.15\\
RegionWise~\cite{ma2022regionwise} & 21.3 & 0.27 & &4.75 & 0.15\\
DeepFill v2~\cite{yu2019free} & 22.1 & 0.28& & 5.20 & 0.16\\
EdgeConnect~\cite{nazeri2019edgeconnect} & 30.5 & 0.28 & & 8.37 & 0.16\\
StableDiffusion*~\cite{rombach2022high} & 8.26 & 0.324 & &2.27 & 0.245 \\
Ours & 6.80 & 0.244 & & 1.03 & 0.141\\
\hline
\end{tabular}
}
\end{center}
\vspace{-1em}
\caption{Comparison of inpainting performance on 30k crops of
size $512\times512$ from test images of Places~\cite{zhou2017places}. Since the high-resolution images of Places is not available, we resize the $256\times256$ images to $512\times512$. The column 40-50\% reports metrics computed over hard examples where 40-50\%
of the image region have to be inpainted. * denotes that the results are reproduced by us.}
\label{tab:inpaintsota}
\end{table}

\section{Experiments}
In order to demonstrate the outstanding application potential of our model, we conduct sufficient experiments based on our base and large models in three different tasks.

\vspace{0.3em}
\noindent\textbf{Implementation details.} In accordance with the training setting used for VQGAN~\cite{esser2021taming} in StableDiffusion~\cite{rombach2022high}, we train our asymmetric VQGAN on the ImageNet~\cite{deng2009imagenet} dataset. During training, we preprocess the image resolution to $256\times 256$, and train our base model for 12 epochs. This process took approximately 5 days on 8 NVIDIA V100 GPUs, with a batch size per GPU of 10 and a learning rate warmed up to 3.6e-4 in the first 5,000 iterations. The learning rate was then decayed with a cosine scheduler. As for our large model, we use 64 NVIDIA V100 GPUs, with a batch size per GPU of 5 and a learning rate warmed up to 7.2e-4 in the first 5,000 iterations. The training for the large model also took around 5 days, with the learning rate decayed using a cosine scheduler too.

\vspace{0.3em}
\noindent\textbf{Evaluation benchmark.} For our inpainting task, we evaluated our model using the same protocol as LaMa~\cite{suvorov2022resolution}, a recent inpainting model, to generate image masks. Our experiments were conducted on two popular datasets: Places~\cite{zhou2017places} and ImageNet~\cite{deng2009imagenet}. Due to the unavailability of high-resolution images in the Places dataset, we resized the $256\times256$ images to $512\times512$. For the ImageNet dataset, we try to randomly select 3 images from each class in the ImageNet validation dataset. However, since some categories do not have 3 images available, we end up selecting a total of 2,968 images for our experiments.

For the paint-by-example task, we evaluate our model using the COCOEE~\cite{yang2022paint} dataset. COCOEE is a COCO exemplar-based image editing benchmark that contains 3,500 source images from the MSCOCO~\cite{lin2014microsoft} validation set. Each image contains only one bounding box, and the mask region is no more than half of the entire image. The corresponding reference image patch is chosen from the MSCOCO training set.

In the text-to-image task, we evaluated our model using the MSCOCO~\cite{lin2014microsoft} validation set. Following the widely-used "Karpathy" split~\cite{Karpathy2017DeepVA}, we used 5,000 images for validation, with each image having approximately 5 captions. Hence, we generated a total of 25,000 images according to the captions. All images were resized to $512\times512$.

\vspace{0.3em}
\noindent\textbf{Evaluation metrics.} 
In the inpainting task, we use FID~\cite{heusel2017gans} and LPIPS~\cite{zhang2018unreasonable} as metrics to evaluate the quality of our model's predictions. Additionally, to showcase the model's ability to preserve the non-edited regions of the image, we report the mean squared error (MSE) between our predictions and ground truth for such non-edited regions. In the paint-by-example task, we measure the quality of our model's output using the mean squared error (MSE) of non-edited image regions and the CLIP score~\cite{radford2021learning}. The CLIP score evaluates the similarity between the edited region and the reference image. Specifically, we resize the two images to $224\times224$, extract their features via CLIP image encoder, and calculate their cosine similarity. A higher CLIP score indicates that the edited region is more similar to the reference image. In the text-to-image task, we evaluate our model's performance using FID and IS~\cite{barratt2018note} as metrics.

\subsection{Evaluation on the Inpainting Task}
\noindent\textbf{Comparison with state-of-the-art methods.} Table~\ref{tab:inpaintsota} shows the comparison of our inpainting approach with other state-of-the-art methods. Our results demonstrate that applying our asymmetric VQGAN to StableDiffusion can improve FID by 1.24. Additionally, our approach outperforms other methods on hard examples where 40-50\% of the image region needs to be inpainted, further highlighting the effectiveness of our conditional branch and larger decoder.

\vspace{0.3em}
\noindent\textbf{Effectiveness of our modules.}
\begin{figure}[t]
\begin{center}
   \includegraphics[width=1\linewidth]{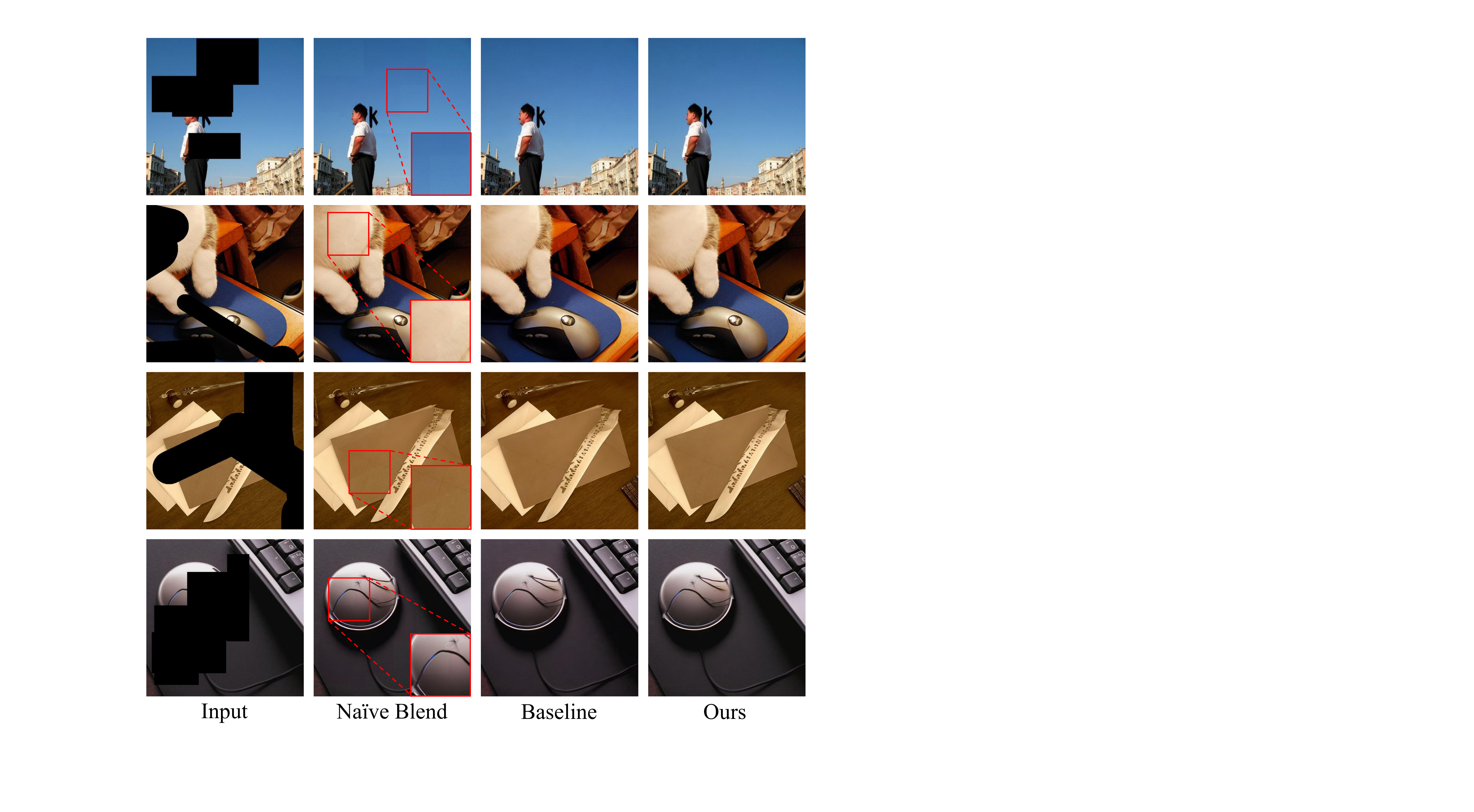}
\end{center}
\vspace{-1em}
   \caption{The harmonization of our asymmetric VQGAN in inpainting task. ``Naive Blend" denotes that adding the non-edited image regions of inputs and the edited regions of results. StableDiffusion~\cite{rombach2022high} is our baseline. All results are reported on ImageNet validation dataset.}
\vspace{-1em}
\label{fig:harm of inpaint}

\end{figure}
This ablation study aims to support the effectiveness of our conditional branch and larger decoder. Table~\ref{tab:inpaint} presents the inpainting results that compare our conditional decoder with the original decoder in StableDiffusion~\cite{rombach2022high}. Specifically, we replace the original decoder with our conditional decoder to decode the quantized vectors obtained from StableDiffusion. It's worth noting that the sample results sent to different decoders are the same since the diffusion process generates various sample results.

Both our base model and larger model show an improvement of approximately 2.00 on FID and 50\% on LPIPS. Furthermore, the error of non-edited image regions is significantly reduced. These results indicate that with the help of the conditional branch, the ability of VQGAN to preserve details has been greatly improved. If we do not use the conditional branch, our base model can still perform the same function as the original decoder. And our larger model can further outperform the original decoder. These results demonstrate that our model is compatible with various applications, whether they have masks or not.
\begin{table}
\begin{center}
\begin{tabular}{l|c|c|c}
\hline
Method & FID$\downarrow$ & LPIPS$\downarrow$ & Pre\_error$\downarrow$ \\
\hline\hline
StableDiffusion~\cite{rombach2022high} & 9.57 & 0.255 & 1082.8$e^-5$ \\
Ours (base) w/ mask & 7.604  & 0.137 & 5.7$e^-5$ \\
Ours (base) w/o mask & 9.48 & 0.248 & 1078.6$e^-5$\\
Ours (large) w/ mask & 7.55  & 0.136 & 2.6$e^-5$ \\
Ours (large) w/o mask & 9.17 & 0.243 & 1047.7$e^-5$\\
\hline
\end{tabular}
\end{center}
\vspace{-1em}
\caption{Results of our different modules in inpainting task. StableDiffusion is our baseline. All results are reported in ImageNet validation dataset.}
\label{tab:inpaint}
\end{table}

\begin{table}
\begin{center}
\begin{tabular}{l|c|c|c}
\hline
Method & FID$\downarrow$ & LPIPS$\downarrow$ & Pre\_error$\downarrow$ \\
\hline\hline
Ours (base) add. & 7.60  & 0.137 & 5.7$e^-5$ \\
Ours (base) concat. & 7.64 & 0.138 & 5.3$e^-5$\\
Ours (large) add. & 7.55  & 0.136 & 2.6$e^-5$ \\
Ours (large) concat. & 7.56 & 0.137 & 4.4$e^-5$\\
\hline
\end{tabular}
\end{center}
\vspace{-1em}
\caption{Ablation study of different blending ways. The ``add." denotes addition and ``concat." denotes concatenation. All results are reported in ImageNet validation dataset.}
\label{tab:blendway}
\end{table}

\begin{table}
\begin{center}
\begin{tabular}{l|c|c}
\hline
Method & CLIP score$\uparrow$ & Pre\_error$\downarrow$ \\
\hline\hline
Blended Diffusion~\cite{avrahami2022blended} & 80.65 & -\\
DCCF~\cite{xue2022dccf} & 82.18 & -\\
StableDiffusion~\cite{rombach2022high}  &  75.33 & -\\
Paint-by-Example~\cite{yang2022paint} & 84.97 & - \\
Paint-by-Example*~\cite{yang2022paint} & 85.67 & 588.86 \\
Ours & 86.35 & 0.76\\
\hline
\end{tabular}
\end{center}
\vspace{-1em}
\caption{Comparison of paint-by-example~\cite{yang2022paint} performance on 3500 images of size $512\times512$. * denotes that the results are reproduced by us. All results are reported on COCOEE.}
\label{tab:pbesota}
\end{table}

\vspace{0.3em}
\noindent\textbf{Addition or concatenation.} 
Our model offers two different blending methods: mask-guided addition and mask-guided concatenation. This ablation study aims to explore how they improve VQGAN. Since the mask is a hard mask, meaning its value is either 0 or 255, our conditional branch does not use Partial Convolutional Layer~\cite{liu2018image} when the blending way is mask-guided addition. This means that the main branch is responsible for the edited regions, while the conditional branch handles the non-edited image regions. As a result, the Partial Convolutional Layer, which infers the information of the edited regions, cannot contribute to the conditional branch. The detailed results are shown in Table~\ref{tab:blendway}. The overall performance of the concatenation is similar to the addition. Finally, we choose the addition as our blending way.

\begin{figure*}[t]
\begin{center}
   \includegraphics[width=1\linewidth]{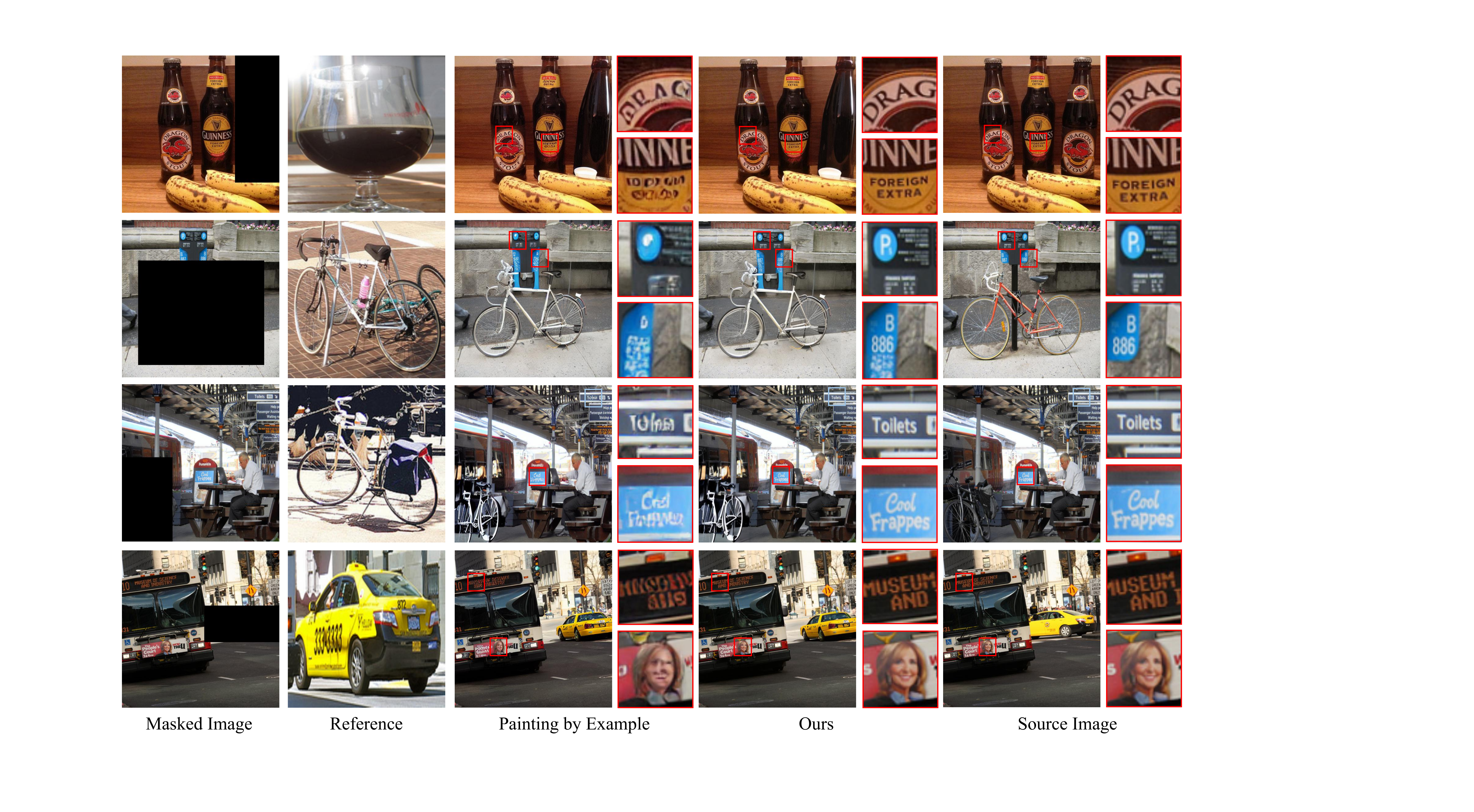}
\end{center}
\vspace{-1em}
   \caption{The preserving ability of our asymmetric VQGAN in paint-by-example~\cite{yang2022paint} task. All results are reported in COCOEE dataset.}
\label{fig:details of pbe}
\vspace{-1.5em}
\end{figure*}

\begin{figure}[!ht]
\begin{center}
   \includegraphics[width=0.95\linewidth]{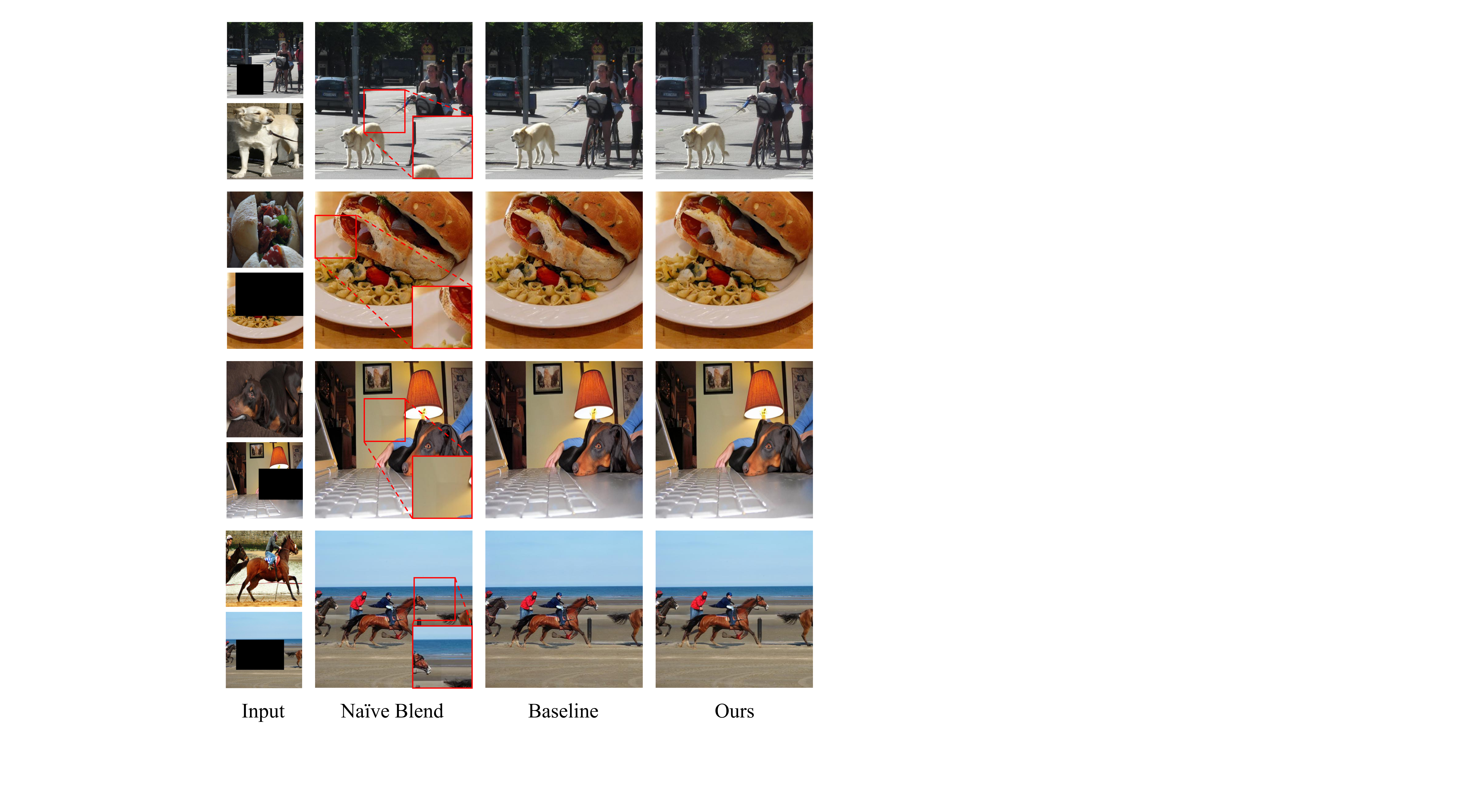}
\end{center}
\vspace{-1em}
   \caption{The harmonization of our asymmetric VQGAN in paint-by-example~\cite{yang2022paint} task. The inputs are masked source images and reference images. `Naive Blend" denotes that adding the non-edited image regions of source images and the edited regions of results. Paint-by-example~\cite{yang2022paint} is our baseline. All results are reported on COCOEE dataset.}
\label{fig:harm of pbe}

\end{figure}

\begin{figure}[!ht]
\begin{center}
   \includegraphics[width=0.95\linewidth]{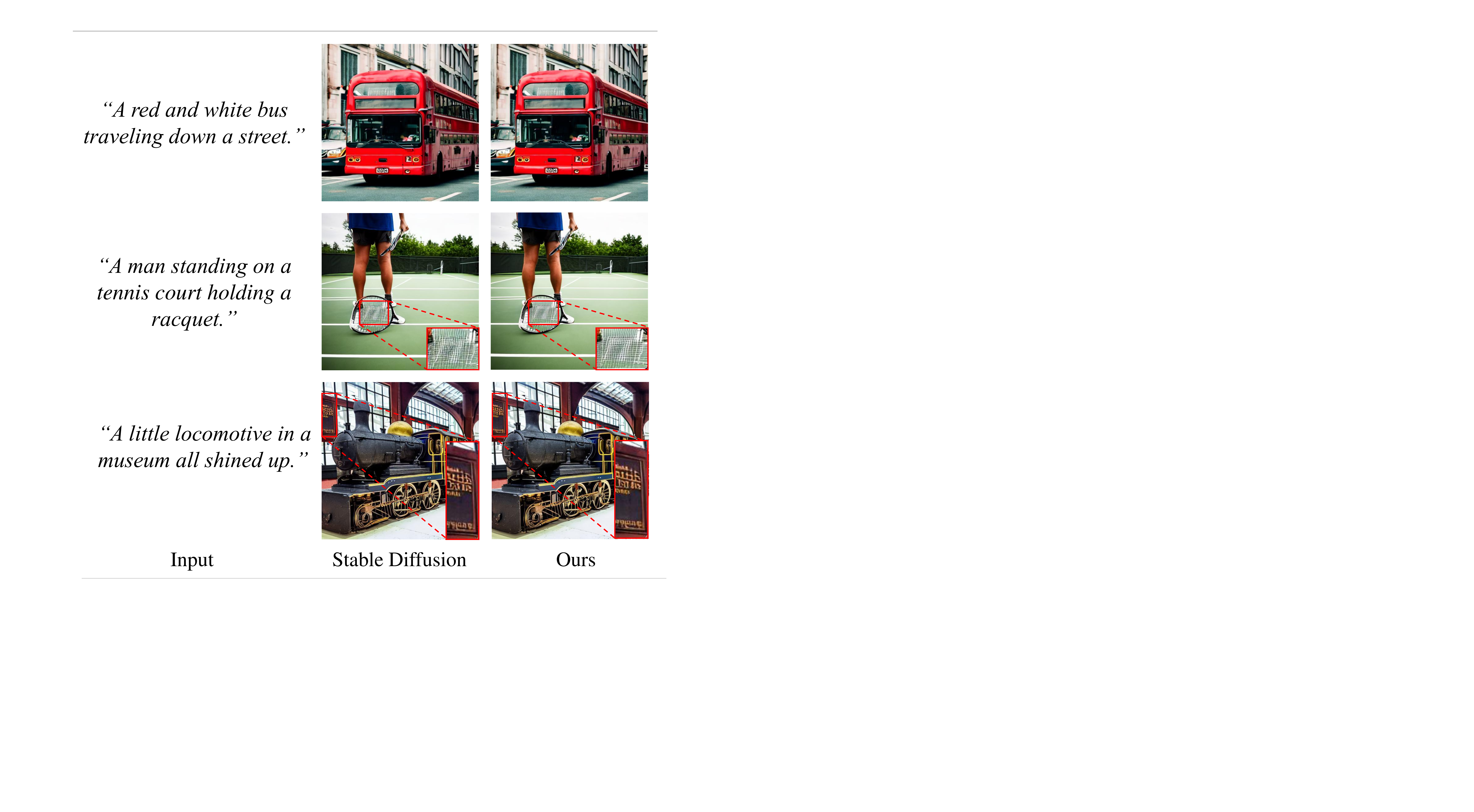}
\end{center}
\vspace{-1em}
   \caption{The comparison between StableDiffusion and ours in text-to-image task. Without the help of our conditional branch, we will get comparable results.}
   \vspace{-2em}
\label{fig:t2i}
\end{figure}

\vspace{0.3em}
\noindent\textbf{Visual Comparison.} 
In order to preserve the details of non-edited image regions, a common approach is to post-process by adding the non-edited regions of the inputs to the edited regions of the results. However, this naive solution can cause non-harmonization issues, as shown in Figure~\ref{fig:harm of inpaint} in the third column. In contrast, our asymmetric VQGAN generates images with the same level of harmonization as the baseline (original StableDiffusion), while preserving many details of the non-edited regions. This indicates that our method effectively preserves details without causing harm to the harmonization of the image. Overall, our results demonstrate that our model with the asymmetric decoder and mask-guided addition blending way can significantly improve the performance of inpainting tasks, while maintaining harmonization and preserving details of non-edited image regions.

\begin{table}
\begin{center}
\begin{tabular}{l|c|c}
\hline
Method & CLIP score$\uparrow$ & Pre\_error$\downarrow$ \\
\hline\hline
Paint-by-Example* & 85.67 & 588.86 \\
Ours (base) & 86.24  & 1.37 \\
Ours (base) w/o mask & 86.15  & 478.47\\
Ours (large) & 86.35 & 0.76\\
Ours (large) w/o mask & 86.32  & 451.67 \\
\hline
\end{tabular}
\end{center}
\vspace{-1em}
\caption{Results of our different modules in paint-by-example~\cite{yang2022paint} task. Paint-by-Example is our baseline. All results are reported in COCOEE dataset.}
\label{tab:pbeab}
\vspace{-1em}
\end{table}

\begin{table}
\begin{center}
\begin{tabular}{l|c|c}
\hline
Method & FID$\downarrow$ & IS$\uparrow$\\
\hline\hline
StableDiffusion~\cite{rombach2022high} & 19.88 & 37.55 \\
Ours (base) w/o mask & 19.92  & 37.52\\
Ours (large) w/o mask & 19.75  & 37.64 \\
\hline
\end{tabular}
\end{center}
\vspace{-1em}
\caption{Results of text-to-image task. StableDiffusion is our baseline. Even though there is no any mask, our asymmetric VQGAN is comparable or superior to the baseline, StableDiffusion. All results are reported on MSCOCO.}
\label{tab:t2iab}
\vspace{-1em}
\end{table}

\subsection{Paint by example} 
Paint-by-example~\cite{yang2022paint} is a novel image editing scenario that semantically alters image content based on an exemplar image. Their approach relies on StableDiffusion as a strong prior, making our model easily applicable to their method.

\vspace{0.3em}
\noindent\textbf{Comparison with state-of-the-art methods.} The comparison results are presented in Table~\ref{tab:pbesota}, where the ``CLIP score" represents the similarity between the edited regions of edited images and reference images, while Pre\_error" denotes the MSE between the non-edited image regions of edited images and source images. Our model can achieve the best performance in both masked and non-edited image regions. It can be seen that our model can achieve the best performance in both masked and non-edited image regions.

\vspace{0.3em}
\noindent\textbf{Effectiveness of our modules.} The purpose of this ablation study is to demonstrate the effectiveness of our conditional branch and larger decoder. The results are presented in Table~\ref{tab:pbeab}. In contrast to the inpainting task, the paint-by-example task involves combining two different images, resulting in a more severe loss of details in non-edited image regions. The non-edited regions of an image can be influenced by the other image, further complicating the task. Our base model reduces the preserving error from 588.86 to 1.37, while our larger model further reduces it to 0.76. Surprisingly, we find that the conditional branch not only improves the preservation of non-edited image regions but also enhances the generation quality of edited regions, as measured by the "CLIP score"

\vspace{0.3em}
\noindent\textbf{Visual results.} We have provided a significant number of visualization results in Figures~\ref{fig:details of pbe} and~\ref{fig:harm of pbe} to demonstrate the preserving ability and harmonization of our asymmetric VQGAN. These visualization results illustrate that our model can be effortlessly applied to various tasks on different datasets while consistently enhancing performance. We firmly believe that our model has immense potential for a wide range of applications.

\subsection{Text-to-Image}
These experiments aim to demonstrate that our asymmetric VQGAN can handle tasks without masks or task-specific priors, in addition to tasks with mask (task-specific priors). The results are presented in Table~\ref{tab:t2iab}. We can observe that when our base model does not use the conditional branch, its performance is comparable to the baseline, which suggests that the training strategy of replacing some masks with a full mask has been successful. Moreover, when our large model does not use the conditional branch, its performance is comparable to the baseline, indicating that the larger decoder can restore more details even without the help of the conditional branch.

\vspace{0.3em}
\noindent\textbf{Visual results.} To support our claim that our asymmetric VQGAN can perform well even without the conditional branch, we present visualization results in Figure~\ref{fig:t2i}. In the first row, it is evident that our model can work effectively without masks and does not produce inferior results. In the second and third rows, we observe that our larger decoder can recover more intricate details to some extent.

\vspace{0.3em}
\noindent\textbf{Effectiveness of our modules.} The purpose of this ablation study is to demonstrate the effectiveness of our conditional branch and larger decoder. The results are presented in Table~\ref{tab:pbeab}. In contrast to the inpainting task, the paint-by-example task involves combining two different images, resulting in a more severe loss of details in non-edited image regions. The non-edited regions of an image can be influenced by the other image, further complicating the task. Our base model reduces the preserving error from 588.86 to 1.37, while our larger model further reduces it to 0.76. Surprisingly, we find that the conditional branch not only improves the preservation of non-edited image regions but also enhances the generation quality of edited regions, as measured by the "CLIP score".

\vspace{0.3em}
\noindent\textbf{Visual results.} We have provided a significant number of visualization results in Figures~\ref{fig:details of pbe} and~\ref{fig:harm of pbe} to demonstrate the preserving ability and harmonization of our asymmetric VQGAN. These visualization results illustrate that our model can be effortlessly applied to various tasks on different datasets while consistently enhancing performance. We firmly believe that our model has immense potential for a wide range of applications.

\subsection{Text-to-Image}
These experiments aim to demonstrate that our asymmetric VQGAN can handle tasks without masks or task-specific priors, in addition to tasks with mask (task-specific priors). The results are presented in Table~\ref{tab:t2iab}. We can observe that when our base model does not use the conditional branch, its performance is comparable to the baseline, which suggests that the training strategy of replacing some masks with a full mask has been successful. Moreover, when our large model does not use the conditional branch, its performance is comparable to the baseline, indicating that the larger decoder can restore more details even without the help of the conditional branch.

\vspace{0.3em}
\noindent\textbf{Visual results.} To support our claim that our asymmetric VQGAN can perform well even without the conditional branch, we present visualization results in Figure~\ref{fig:t2i}. In the first row, it is evident that our model can work effectively without masks and does not produce inferior results. In the second and third rows, we observe that our larger decoder can recover more intricate details to some extent.

\section{Conclusion}
In this paper, we present a novel asymmetric VQGAN for StableDiffusion with two new design features. Firstly, our decoder incorporates an additional conditional branch, allowing it to accept both the output of the VQGAN encoder and task-specific priors as input. Secondly, our decoder is designed to be more complex (e.g. deeper and wider) than the encoder, enabling it to better preserve local details of non-edited regions and recover details from the quantized output of the encoder. Our asymmetric VQGAN architecture is highly efficient for both training and inference, and can be used for local editing tasks and pure text-to-image generation tasks. Through extensive experimentation on two representative tasks, we demonstrate the effectiveness of our asymmetric VQGAN design. Moving forward, we plan to explore whether scaling up the decoder could further improve the quality of our results.

{\small
\bibliographystyle{ieee_fullname}
\bibliography{egbib}

\begin{thebibliography}{10}\itemsep=-1pt

\bibitem{avrahami2022blended}
Omri Avrahami, Dani Lischinski, and Ohad Fried.
\newblock Blended diffusion for text-driven editing of natural images.
\newblock In {\em CVPR}, pages 18208--18218, 2022.

\bibitem{balaji2022ediffi}
Yogesh Balaji, Seungjun Nah, Xun Huang, Arash Vahdat, Jiaming Song, Karsten
  Kreis, Miika Aittala, Timo Aila, Samuli Laine, Bryan Catanzaro, et~al.
\newblock ediffi: Text-to-image diffusion models with an ensemble of expert
  denoisers.
\newblock {\em arXiv preprint arXiv:2211.01324}, 2022.

\bibitem{barratt2018note}
Shane Barratt and Rishi Sharma.
\newblock A note on the inception score.
\newblock {\em arXiv preprint arXiv:1801.01973}, 2018.

\bibitem{batzolis2021conditional}
Georgios Batzolis, Jan Stanczuk, Carola-Bibiane Sch{\"o}nlieb, and Christian
  Etmann.
\newblock Conditional image generation with score-based diffusion models.
\newblock {\em arXiv preprint arXiv:2111.13606}, 2021.

\bibitem{brooks2022instructpix2pix}
Tim Brooks, Aleksander Holynski, and Alexei~A Efros.
\newblock Instructpix2pix: Learning to follow image editing instructions.
\newblock {\em arXiv preprint arXiv:2211.09800}, 2022.

\bibitem{choi2021ilvr}
Jooyoung Choi, Sungwon Kim, Yonghyun Jeong, Youngjune Gwon, and Sungroh Yoon.
\newblock Ilvr: Conditioning method for denoising diffusion probabilistic
  models.
\newblock {\em arXiv preprint arXiv:2108.02938}, 2021.

\bibitem{deng2009imagenet}
Jia Deng, Wei Dong, Richard Socher, Li-Jia Li, Kai Li, and Li Fei-Fei.
\newblock Imagenet: A large-scale hierarchical image database.
\newblock In {\em CVPR}, pages 248--255. Ieee, 2009.

\bibitem{dhariwal2021diffusion}
Prafulla Dhariwal and Alexander Nichol.
\newblock Diffusion models beat gans on image synthesis.
\newblock {\em Advances in Neural Information Processing Systems},
  34:8780--8794, 2021.

\bibitem{esser2021imagebart}
Patrick Esser, Robin Rombach, Andreas Blattmann, and Bjorn Ommer.
\newblock Imagebart: Bidirectional context with multinomial diffusion for
  autoregressive image synthesis.
\newblock {\em Advances in Neural Information Processing Systems},
  34:3518--3532, 2021.

\bibitem{esser2021taming}
Patrick Esser, Robin Rombach, and Bjorn Ommer.
\newblock Taming transformers for high-resolution image synthesis.
\newblock In {\em CVPR}, pages 12873--12883, 2021.

\bibitem{gu2022vector}
Shuyang Gu, Dong Chen, Jianmin Bao, Fang Wen, Bo Zhang, Dongdong Chen, Lu Yuan,
  and Baining Guo.
\newblock Vector quantized diffusion model for text-to-image synthesis.
\newblock In {\em CVPR}, pages 10696--10706, 2022.

\bibitem{heusel2017gans}
Martin Heusel, Hubert Ramsauer, Thomas Unterthiner, Bernhard Nessler, and Sepp
  Hochreiter.
\newblock Gans trained by a two time-scale update rule converge to a local nash
  equilibrium.
\newblock {\em Advances in neural information processing systems}, 30, 2017.

\bibitem{ho2020denoising}
Jonathan Ho, Ajay Jain, and Pieter Abbeel.
\newblock Denoising diffusion probabilistic models.
\newblock {\em Advances in Neural Information Processing Systems},
  33:6840--6851, 2020.

\bibitem{ho2022cascaded}
Jonathan Ho, Chitwan Saharia, William Chan, David~J Fleet, Mohammad Norouzi,
  and Tim Salimans.
\newblock Cascaded diffusion models for high fidelity image generation.
\newblock {\em J. Mach. Learn. Res.}, 23(47):1--33, 2022.

\bibitem{ho2022video}
Jonathan Ho, Tim Salimans, Alexey Gritsenko, William Chan, Mohammad Norouzi,
  and David~J Fleet.
\newblock Video diffusion models.
\newblock {\em arXiv preprint arXiv:2204.03458}, 2022.

\bibitem{isola2017image}
Phillip Isola, Jun-Yan Zhu, Tinghui Zhou, and Alexei~A Efros.
\newblock Image-to-image translation with conditional adversarial networks.
\newblock In {\em CVPR}, 2017.

\bibitem{Karpathy2017DeepVA}
Andrej Karpathy and Li Fei-Fei.
\newblock Deep visual-semantic alignments for generating image descriptions.
\newblock {\em IEEE Transactions on Pattern Analysis and Machine Intelligence},
  39:664--676, 2017.

\bibitem{kong2020diffwave}
Zhifeng Kong, Wei Ping, Jiaji Huang, Kexin Zhao, and Bryan Catanzaro.
\newblock Diffwave: A versatile diffusion model for audio synthesis.
\newblock {\em arXiv preprint arXiv:2009.09761}, 2020.

\bibitem{li2022diffusion}
Xiang~Lisa Li, John Thickstun, Ishaan Gulrajani, Percy Liang, and Tatsunori~B
  Hashimoto.
\newblock Diffusion-lm improves controllable text generation.
\newblock {\em arXiv preprint arXiv:2205.14217}, 2022.

\bibitem{lin2014microsoft}
Tsung-Yi Lin, Michael Maire, Serge Belongie, James Hays, Pietro Perona, Deva
  Ramanan, Piotr Doll{\'a}r, and C~Lawrence Zitnick.
\newblock Microsoft coco: Common objects in context.
\newblock In {\em ECCV}, pages 740--755, 2014.

\bibitem{liu2018image}
Guilin Liu, Fitsum~A Reda, Kevin~J Shih, Ting-Chun Wang, Andrew Tao, and Bryan
  Catanzaro.
\newblock Image inpainting for irregular holes using partial convolutions.
\newblock In {\em ECCV}, pages 85--100, 2018.

\bibitem{liu2022reduce}
Qiankun Liu, Zhentao Tan, Dongdong Chen, Qi Chu, Xiyang Dai, Yinpeng Chen,
  Mengchen Liu, Lu Yuan, and Nenghai Yu.
\newblock Reduce information loss in transformers for pluralistic image
  inpainting.
\newblock In {\em Proceedings of the IEEE/CVF Conference on Computer Vision and
  Pattern Recognition}, pages 11347--11357, 2022.

\bibitem{ma2022regionwise}
Yuqing Ma, Xianglong Liu, Shihao Bai, Lei Wang, Aishan Liu, Dacheng Tao, and
  Edwin~R Hancock.
\newblock Regionwise generative adversarial image inpainting for large missing
  areas.
\newblock {\em IEEE Transactions on Cybernetics}, 2022.

\bibitem{mokady2022null}
Ron Mokady, Amir Hertz, Kfir Aberman, Yael Pritch, and Daniel Cohen-Or.
\newblock Null-text inversion for editing real images using guided diffusion
  models.
\newblock {\em arXiv preprint arXiv:2211.09794}, 2022.

\bibitem{nazeri2019edgeconnect}
Kamyar Nazeri, Eric Ng, Tony Joseph, Faisal~Z Qureshi, and Mehran Ebrahimi.
\newblock Edgeconnect: Generative image inpainting with adversarial edge
  learning.
\newblock {\em arXiv preprint arXiv:1901.00212}, 2019.

\bibitem{nichol2021glide}
Alex Nichol, Prafulla Dhariwal, Aditya Ramesh, Pranav Shyam, Pamela Mishkin,
  Bob McGrew, Ilya Sutskever, and Mark Chen.
\newblock Glide: Towards photorealistic image generation and editing with
  text-guided diffusion models.
\newblock {\em arXiv preprint arXiv:2112.10741}, 2021.

\bibitem{peng2021generating}
Jialun Peng, Dong Liu, Songcen Xu, and Houqiang Li.
\newblock Generating diverse structure for image inpainting with hierarchical
  vq-vae.
\newblock In {\em CVPR}, pages 10775--10784, 2021.

\bibitem{radford2021learning}
Alec Radford, Jong~Wook Kim, Chris Hallacy, Aditya Ramesh, Gabriel Goh,
  Sandhini Agarwal, Girish Sastry, Amanda Askell, Pamela Mishkin, Jack Clark,
  et~al.
\newblock Learning transferable visual models from natural language
  supervision.
\newblock In {\em ICML}, pages 8748--8763, 2021.

\bibitem{ramesh2022hierarchical}
Aditya Ramesh, Prafulla Dhariwal, Alex Nichol, Casey Chu, and Mark Chen.
\newblock Hierarchical text-conditional image generation with clip latents.
\newblock {\em arXiv preprint arXiv:2204.06125}, 2022.

\bibitem{razavi2019generating}
Ali Razavi, Aaron Van~den Oord, and Oriol Vinyals.
\newblock Generating diverse high-fidelity images with vq-vae-2.
\newblock {\em Advances in neural information processing systems}, 32, 2019.

\bibitem{rombach2022high}
Robin Rombach, Andreas Blattmann, Dominik Lorenz, Patrick Esser, and Bj{\"o}rn
  Ommer.
\newblock High-resolution image synthesis with latent diffusion models.
\newblock In {\em CVPR}, pages 10684--10695, 2022.

\bibitem{saharia2022photorealistic}
Chitwan Saharia, William Chan, Saurabh Saxena, Lala Li, Jay Whang, Emily
  Denton, Seyed Kamyar~Seyed Ghasemipour, Burcu~Karagol Ayan, S~Sara Mahdavi,
  Rapha~Gontijo Lopes, et~al.
\newblock Photorealistic text-to-image diffusion models with deep language
  understanding.
\newblock {\em arXiv preprint arXiv:2205.11487}, 2022.

\bibitem{simonyan2014very}
Karen Simonyan and Andrew Zisserman.
\newblock Very deep convolutional networks for large-scale image recognition.
\newblock {\em arXiv preprint arXiv:1409.1556}, 2014.

\bibitem{sohl2015deep}
Jascha Sohl-Dickstein, Eric Weiss, Niru Maheswaranathan, and Surya Ganguli.
\newblock Deep unsupervised learning using nonequilibrium thermodynamics.
\newblock In {\em International Conference on Machine Learning}, pages
  2256--2265. PMLR, 2015.

\bibitem{song2020denoising}
Jiaming Song, Chenlin Meng, and Stefano Ermon.
\newblock Denoising diffusion implicit models.
\newblock {\em arXiv preprint arXiv:2010.02502}, 2020.

\bibitem{suvorov2022resolution}
Roman Suvorov, Elizaveta Logacheva, Anton Mashikhin, Anastasia Remizova,
  Arsenii Ashukha, Aleksei Silvestrov, Naejin Kong, Harshith Goka, Kiwoong
  Park, and Victor Lempitsky.
\newblock Resolution-robust large mask inpainting with fourier convolutions.
\newblock In {\em WACV}, pages 2149--2159, 2022.

\bibitem{wang2022semantic}
Weilun Wang, Jianmin Bao, Wengang Zhou, Dongdong Chen, Dong Chen, Lu Yuan, and
  Houqiang Li.
\newblock Semantic image synthesis via diffusion models.
\newblock {\em arXiv preprint arXiv:2207.00050}, 2022.

\bibitem{wang2022sindiffusion}
Weilun Wang, Jianmin Bao, Wengang Zhou, Dongdong Chen, Dong Chen, Lu Yuan, and
  Houqiang Li.
\newblock Sindiffusion: Learning a diffusion model from a single natural image.
\newblock {\em arXiv preprint arXiv:2211.12445}, 2022.

\bibitem{xue2022dccf}
Ben Xue, Shenghui Ran, Quan Chen, Rongfei Jia, Binqiang Zhao, and Xing Tang.
\newblock Dccf: Deep comprehensible color filter learning framework for
  high-resolution image harmonization.
\newblock In {\em ECCV}, pages 300--316, 2022.

\bibitem{yang2022paint}
Binxin Yang, Shuyang Gu, Bo Zhang, Ting Zhang, Xuejin Chen, Xiaoyan Sun, Dong
  Chen, and Fang Wen.
\newblock Paint by example: Exemplar-based image editing with diffusion models.
\newblock {\em arXiv preprint arXiv:2211.13227}, 2022.

\bibitem{yu2019free}
Jiahui Yu, Zhe Lin, Jimei Yang, Xiaohui Shen, Xin Lu, and Thomas~S Huang.
\newblock Free-form image inpainting with gated convolution.
\newblock In {\em ICCV}, pages 4471--4480, 2019.

\bibitem{zhang2018unreasonable}
Richard Zhang, Phillip Isola, Alexei~A Efros, Eli Shechtman, and Oliver Wang.
\newblock The unreasonable effectiveness of deep features as a perceptual
  metric.
\newblock In {\em CVPR}, pages 586--595, 2018.

\bibitem{zhao2021large}
Shengyu Zhao, Jonathan Cui, Yilun Sheng, Yue Dong, Xiao Liang, Eric~I Chang,
  and Yan Xu.
\newblock Large scale image completion via co-modulated generative adversarial
  networks.
\newblock {\em arXiv preprint arXiv:2103.10428}, 2021.

\bibitem{zhou2017places}
Bolei Zhou, Agata Lapedriza, Aditya Khosla, Aude Oliva, and Antonio Torralba.
\newblock Places: A 10 million image database for scene recognition.
\newblock {\em IEEE transactions on pattern analysis and machine intelligence},
  40(6):1452--1464, 2017.

\end{thebibliography}
}

\appendix
\twocolumn[\centering\section*{Appendices}]
In the following section, we first introduce the loss objectives for Asymmetric VQGAN, then we present the other technique for compressing image space to latent space KL-reg, which shares a similar spirit to VQGAN. Finally, we introduce the architecture of our models.
\section{Training Objectives}
For the training of Asymmetric VQGAN, we fix the weights of the encoder and the codebook and employ a reconstruction and adversarial loss to train the decoder. The reconstruction loss is the sum of pixel-loss and perceptual loss. Pixel-level loss is the MAE loss used between each pixel of the output image from the quantized vector $\mathbf{z}$ and input image $\mathbf{x}$, which can be denoted by $\mathcal{L}_{pixel}=\frac{1}{3HW}|\mathbf{x}-\hat{\mathbf{x}}|$, where  $\hat{\mathbf{x}} = Dec(\mathbf{z})$ denotes the output image, $H$ and $w$ denote the high and width of the image, respectively. Moreover, the perceptual loss $\mathcal{L}_{percep}$ is calculated with VGG16~\cite{simonyan2014very} network, and it can be formulated as:
\begin{equation}
\begin{aligned}
\mathcal{L}_{percep}^k(\mathbf{x},\hat{\mathbf{x}})&= \frac{1}{H^kW^k}\text{f}^k(\|\phi^k(\mathbf{x}) - \phi^k(\mathbf{\hat{\mathbf{x}}})\|^2),\\
\mathcal{L}_{percep}&=\sum\limits^4_{k=0}{\mathcal{L}_{percep}^k(\mathbf{x},\hat{\mathbf{x}})}
\end{aligned}
\end{equation}
where $k=\{0,1,2,3,4\}$, $H^k$ and $W^k$ denote the high and width of the image feature in $k$-th layer, respectively. $\text{f}^k$ means a convolution operation with $1\times1$ kernel to reduce the channel to 1. And $\phi^k(\cdot)$ is the $k$-th layer of pretrained VGG16 network. 

To sum up, the reconstruction loss can be generalized as follows:
\begin{equation}
\begin{aligned}
\mathcal{L}_{\mathrm{REC}}(Dec)&=\mathcal{L}_{pixel} + \mathcal{L}_{percep}.
\end{aligned}
\end{equation}

Another important loss to improve the quality of the generated results is the GAN loss, and the equation is defined by:
\begin{equation}
\begin{aligned}
\mathcal{L}_{\mathrm{GAN}}(Dec,D)&= \mathop{\min}_{Dec}\mathop{\max}_{D}E_{x}[log(D(\mathbf{x}))]  \\
&+ E_{\hat{\mathbf{x}}}[log(1-D(\hat{\mathbf{x}})],
\end{aligned}
\end{equation}
where $D$ denotes a patch-based discriminator~\cite{isola2017image}, $Dec$ denotes the generator which is our asymmetric decoder.

Therefore the overall objective for training the decoder model $\mathcal{L}$ then reads
\begin{equation}
\begin{aligned}
\mathcal{L} = \mathcal{L}_{\mathrm{REC}}( Dec) + \lambda \mathcal{L}_{\mathrm{GAN}}(Dec,D),
\end{aligned}
\end{equation}
where we compute the adaptive weight $\lambda$ according to
\begin{equation}
\begin{aligned}
\lambda=\frac{\nabla_{G_L}\left[\mathcal{L}_{\mathrm{pixel}}\right]}{\nabla_{G_L}\left[\mathcal{L}_{\mathrm{GAN}}\right]+\delta}.
\end{aligned}
\label{eqn:gan_loss}
\end{equation}
The $\nabla_{G_L}[\cdot]$ denotes the gradient of its input w.r.t. the last layer $L$ of our decoder and $\delta = 10^{-4}$ is used for numerical stability.

\section{KL-reg for Training VAEGAN}
Besides VQGAN, in some versions of StableDiffusion, the compressing from pixel space to latent space is trained by a KL-reg, which can be regarded as VAEGAN. The VAEGAN shares a similar purpose to VQGAN since they all try to avoid arbitrarily high-variance latent spaces.

For VAEGAN, the encoder network outputs the mean and covariance of the latent vector, \textit{i.e.}, $\mu$ and $\epsilon$. $\mathcal{L}_{kl}$ is used to reduce the gap between the prior and the proposal distributions.
\begin{equation}
\begin{aligned}
\mathcal{L}_{kl}=\frac{1}{2}\left(\boldsymbol{\mu}^T \boldsymbol{\mu}+\operatorname{sum}(\exp (\boldsymbol{\epsilon})-\boldsymbol{\epsilon}-1)\right).
\end{aligned}
\end{equation}

and the VAEGAN loss can be formulated as:
\begin{equation}
\begin{aligned}
\mathcal{L}_{\mathrm{VAEGAN}}(\{Enc, Dec\})&=\mathcal{L}_{pixel} + \mathcal{L}_{percep} +  \mathcal{L}_{kl}.
\end{aligned}
\end{equation}
We also include a GAN loss(Eqn.~\ref{eqn:gan_loss}) to improve the quality. For the training of the asymmetric VAEGAN, the encoder part is also fixed, and we only apply the  $\mathcal{L}_{pixel} + \mathcal{L}_{percep} + \mathcal{L}_{\mathrm{GAN}}(Dec,D)$ loss functions to update the decoder. The $\mathcal{L}_{kl}$ is omitted during the training process.

\section{Architecture of Our Models}
We further present the details of the architecture of our base model in Table~\ref{tab:basearch}, and the architecture of our large model is shown in Table~\ref{tab:largearch}.

\begin{table}[t]
\centering
\begin{center}
\scalebox{0.9}{
\begin{tabular}{c|c|c|c}

\hline
Branch & $l$-th & Layer / kernel & Output Size \\
\hline

\multirow{5}{*}{Condition} & 0 &PConv2d / (3, 3)& 128 $\times$ 512 $\times$ 512\\ 
\cline{2-4}
\multirow{5}{*}{} & 1 & PConv2d / (3, 3) & 256 $\times$ 512 $\times$ 512\\ 
\cline{2-4}
\multirow{5}{*}{} & 2 & PConv2d / (4, 4) & 512 $\times$ 256 $\times$ 256\\ 
\cline{2-4}
\multirow{5}{*}{} & 3 &PConv2d / (4, 4)	& 512 $\times$ 128 $\times$ 128\\ 
\cline{2-4}
\multirow{5}{*}{} & 4 & PConv2d / (4, 4)	& 512 $\times$ 64 $\times$ 64\\ 
\hline

\multirow{23}{*}{Main} &\multirow{4}{*}{Out} &	Concat & 257 $\times$ 512 $\times$ 512\\ 
\multirow{23}{*}{} &\multirow{4}{*}{} &	Conv2d / (1, 1) & 128 $\times$ 512 $\times$ 512\\ 
\multirow{23}{*}{} &\multirow{4}{*}{} &	\demph{GroupNorm / (1, 1)} & 128 $\times$ 512 $\times$ 512\\
\multirow{23}{*}{} &\multirow{4}{*}{} &	\demph{Conv2d / (3, 3)}& 3 $\times$ 512 $\times$ 512\\
\cline{2-4}
\multirow{23}{*}{} &\multirow{3}{*}{0} &	Concat & 513 $\times$ 512 $\times$ 512\\ 
\multirow{23}{*}{} &\multirow{3}{*}{} &	Conv2d / (1, 1) & 256 $\times$ 512 $\times$ 512\\ 
\multirow{23}{*}{} & \multirow{3}{*}{} & \demph{ResBlock $\times$3} & 128 $\times$ 512 $\times$ 512\\ 
\cline{2-4}
\multirow{23}{*}{} &\multirow{4}{*}{1} &	Concat & 1025 $\times$ 256 $\times$ 256\\ 
\multirow{23}{*}{} &\multirow{4}{*}{} &	Conv2d / (1, 1) & 512 $\times$ 256 $\times$ 256\\ 
\multirow{23}{*}{} & \multirow{4}{*}{} & \demph{ResBlock $\times$3} & 256 $\times$ 256 $\times$ 256\\ 
\multirow{23}{*}{} & \multirow{4}{*}{} & \demph{Upsample} & 256 $\times$ 512 $\times$ 512\\
\cline{2-4}
\multirow{23}{*}{} &\multirow{4}{*}{2} &	Concat & 1025 $\times$ 128 $\times$ 128\\ 
\multirow{23}{*}{} &\multirow{4}{*}{} &	Conv2d / (1, 1) & 512 $\times$ 128 $\times$ 128\\ 
\multirow{23}{*}{} & \multirow{4}{*}{} & \demph{ResBlock $\times$3} & 512 $\times$ 128 $\times$ 128\\ 
\multirow{23}{*}{} & \multirow{4}{*}{} & \demph{Upsample} & 512 $\times$ 256 $\times$ 256\\
\cline{2-4}
\multirow{23}{*}{} &\multirow{4}{*}{3} &	Concat & 1025 $\times$ 64 $\times$ 64\\ 
\multirow{23}{*}{} &\multirow{4}{*}{} &	Conv2d / (1, 1) & 512 $\times$ 64 $\times$ 64\\ 
\multirow{23}{*}{} & \multirow{4}{*}{} & \demph{ResBlock $\times$3} & 512 $\times$ 64 $\times$ 64\\ 
\multirow{23}{*}{} & \multirow{4}{*}{} & \demph{Upsample} & 512 $\times$ 128 $\times$ 128\\
\cline{2-4}
\multirow{23}{*}{} & \multirow{4}{*}{4} & \demph{Conv2d / (3, 3)}& \multirow{4}{*}{512 $\times$ 64 $\times$ 64}\\ 
\multirow{23}{*} & \multirow{4}{*}{} & \demph{ResBlock} & \multirow{4}{*}{} \\ 
\multirow{23}{*}{} & \multirow{4}{*}{} & \demph{AttnBlock}	& \multirow{4}{*}{}\\ 
\multirow{23}{*}{} &\multirow{4}{*}{} & \demph{ResBlock}	& \multirow{4}{*}{}\\ 
\hline

\end{tabular}
}
\end{center}
\caption{Architecture of our base model. The way of mask guided blend is concatenation. ``PConv" denotes Partial Convolutional Layer~\cite{liu2018image}. The \demph{gray font} denotes the  vanilla decoder.}
\label{tab:basearch}
\end{table}

\begin{table}[t]
\centering
\begin{center}
\scalebox{0.9}{
\begin{tabular}{c|c|c|c}

\hline
Branch & $l$-th & Layer / kernel & Output Size \\
\hline

\multirow{5}{*}{Condition} & 0 &PConv2d / (3, 3)& 192 $\times$ 512 $\times$ 512\\ 
\cline{2-4}
\multirow{5}{*}{} & 1 & PConv2d / (3, 3) & 384 $\times$ 512 $\times$ 512\\ 
\cline{2-4}
\multirow{5}{*}{} & 2 & PConv2d / (4, 4) & 768 $\times$ 256 $\times$ 256\\ 
\cline{2-4}
\multirow{5}{*}{} & 3 &PConv2d / (4, 4)	& 768 $\times$ 128 $\times$ 128\\ 
\cline{2-4}
\multirow{5}{*}{} & 4 & PConv2d / (4, 4)	& 768 $\times$ 64 $\times$ 64\\ 
\hline

\multirow{23}{*}{Main} &\multirow{4}{*}{Out} &	Concat & 385 $\times$ 512 $\times$ 512\\ 
\multirow{23}{*}{} &\multirow{4}{*}{} &	Conv2d / (1, 1) & 192 $\times$ 512 $\times$ 512\\ 
\multirow{23}{*}{} &\multirow{4}{*}{} &	GroupNorm / (1, 1) & 192 $\times$ 512 $\times$ 512\\
\multirow{23}{*}{} &\multirow{4}{*}{} &	Conv2d / (3, 3) & 3 $\times$ 512 $\times$ 512\\
\cline{2-4}
\multirow{23}{*}{} &\multirow{3}{*}{0} &	Concat & 769 $\times$ 512 $\times$ 512\\ 
\multirow{23}{*}{} &\multirow{3}{*}{} &	Conv2d / (1, 1) & 384 $\times$ 512 $\times$ 512\\ 
\multirow{23}{*}{} & \multirow{3}{*}{} & ResBlock $\times$4 & 192 $\times$ 512 $\times$ 512\\ 
\cline{2-4}
\multirow{23}{*}{} &\multirow{4}{*}{1} &	Concat & 1537 $\times$ 256 $\times$ 256\\ 
\multirow{23}{*}{} &\multirow{4}{*}{} &	Conv2d / (1, 1) & 768 $\times$ 256 $\times$ 256\\ 
\multirow{23}{*}{} & \multirow{4}{*}{} & ResBlock $\times$4 & 384 $\times$ 256 $\times$ 256\\ 
\multirow{23}{*}{} & \multirow{4}{*}{} & Upsample & 384 $\times$ 512 $\times$ 512\\
\cline{2-4}
\multirow{23}{*}{} &\multirow{4}{*}{2} &	Concat & 1537 $\times$ 128 $\times$ 128\\ 
\multirow{23}{*}{} &\multirow{4}{*}{} &	Conv2d / (1, 1) & 768 $\times$ 128 $\times$ 128\\ 
\multirow{23}{*}{} & \multirow{4}{*}{} & ResBlock $\times$4 & 768 $\times$ 128 $\times$ 128\\ 
\multirow{23}{*}{} & \multirow{4}{*}{} & Upsample & 768 $\times$ 256 $\times$ 256\\
\cline{2-4}
\multirow{23}{*}{} &\multirow{4}{*}{3} &	Concat & 1537 $\times$ 64 $\times$ 64\\ 
\multirow{23}{*}{} &\multirow{4}{*}{} &	Conv2d / (1, 1) & 768 $\times$ 64 $\times$ 64\\ 
\multirow{23}{*}{} & \multirow{4}{*}{} & ResBlock $\times$4 & 768 $\times$ 64 $\times$ 64\\ 
\multirow{23}{*}{} & \multirow{4}{*}{} & Upsample & 768 $\times$ 128 $\times$ 128\\
\cline{2-4}
\multirow{23}{*}{} & \multirow{4}{*}{4} & Conv2d / (3, 3)& \multirow{4}{*}{768 $\times$ 64 $\times$ 64}\\ 
\multirow{23}{*} & \multirow{4}{*}{} & ResBlock	& \multirow{4}{*}{} \\ 
\multirow{23}{*}{} & \multirow{4}{*}{} & AttnBlock	& \multirow{4}{*}{}\\ 
\multirow{23}{*}{} &\multirow{4}{*}{} & ResBlock	& \multirow{4}{*}{}\\ 
\hline

\end{tabular}
}
\end{center}
\caption{Architecture of our large model. The way of mask guided blend is concatenation. ``PConv" denotes Partial Convolutional Layer~\cite{liu2018image}.}
\label{tab:largearch}
\end{table}

\section{Larger Decoder}

In contrast to the conventional balanced size between encoder and decoder, we propose to design the decoder to be heavier than the encoder (in-balanced design). This is based on the important observation that the main computation bottleneck of StableDiffusion~\cite{rombach2022high} lies in the diffusion process but not VQGAN. This design can not only improve the quality for both masked and unmasked regions for local editing, but benefit the pure text-to-image generation performance. 

As shown in Table~\ref{tab:large-in} and Table~\ref{tab:large-t2i}, our method not only benefits the masked region generation quality in local editing (FID and LPIPS improvement for the whole edited image) but also improves the original text-to-image generation task quality upon StableDiffusion, whereas all the blending methods cannot achieve this goal.
\begin{table}[h]
\begin{center}
\begin{tabular}{l|c|c|c}
\hline
Method & FID$\downarrow$ & LPIPS$\downarrow$ & Pre\_error$\downarrow$ \\
\hline\hline
Base decoder & 7.60  & 0.137 & 5.7$e^-5$ \\
Large1 decoder & 7.55  & 0.136 & 2.6$e^-5$ \\
Large$\times$2 decoder & 7.49 & 0.134 & 2.1$e^-5$\\
\hline
\end{tabular}
\end{center}
\caption{Effectiveness of our larger decoder in inpainting task. ``Large decoder" denotes we increase the width and depth of the decoder by 1.5 times. ``Large$\times$2 decoder" denotes we increase the width of the decoder by 2 times and the depth by 2.5 times.}
\label{tab:large-in}
\end{table}

\begin{table}[h]
\begin{center}
\begin{tabular}{l|c|c}
\hline
Method & FID$\downarrow$ & IS$\uparrow$\\
\hline\hline
Base decoder w/o mask & 19.92  & 37.52\\
Large decoder w/o mask & 19.75  & 37.64 \\
Large$\times$2 decoder w/o mask & 19.68  & 37.74 \\
\hline
\end{tabular}
\end{center}
\caption{Effectiveness of our larger decoder on text-to-image task.}
\label{tab:large-t2i}
\end{table}

\end{document}